\documentclass{article}
\usepackage{ijcai20}

\usepackage{times}

\usepackage{soul}
\usepackage{url}
\usepackage[hidelinks]{hyperref}
\usepackage[utf8]{inputenc}
\usepackage[small]{caption}
\usepackage{graphicx}
\usepackage{amsmath}
\usepackage{amsthm}
\usepackage{array}
\usepackage{color}
\usepackage{booktabs}
\usepackage{multirow}
\usepackage{algorithm}
\usepackage{algorithmic}
\title{Continual Local Replacement for Few-shot Learning}

\author{
Canyu Le$^1$\and
Zhonggui Chen$^1$\footnote{Corresponding Author}\and
Xihan Wei$^2$\and
Biao Wang$^2$\and
Lei Zhang$^2$\\
\affiliations
$^1$School of Informatics, Xiamen University\\
$^2$DAMO Academy, Alibaba Group\\
\emails
lecanyu@gmail.com,
chenzhonggui@xmu.edu.cn, \\
\{xihan.wxh, wb.wangbiao, lei.zhang.lz\}@alibaba-inc.com
}

\begin{document}

\maketitle

\begin{abstract}
The goal of few-shot learning is to learn a model that can recognize novel classes based on one or few training data.
It is challenging mainly due to two aspects:
(1) it lacks good feature representation of novel classes;
(2) a few of labeled data could not accurately represent the true data distribution and thus it's hard to learn a good decision function for classification.
In this work, we use a sophisticated network architecture to learn better feature representation and focus on the second issue.
A novel continual local replacement strategy is proposed to address the data deficiency problem.
It takes advantage of the content in unlabeled images to \emph{continually} enhance labeled ones.
Specifically, a pseudo labeling method is adopted to constantly select semantically similar images on the fly.
Original labeled images will be locally replaced by the selected images for the next epoch training.
In this way, the model can directly learn new semantic information from unlabeled images and the capacity of supervised signals in the embedding space can be significantly enlarged.
This allows the model to improve generalization and learn a better decision boundary for classification.
Our method is conceptually simple and easy to implement.
Extensive experiments demonstrate that it can achieve state-of-the-art results on various few-shot image recognition benchmarks.
\end{abstract}

\section{Introduction}
While deep learning has achieved remarkable results in image recognition~\cite{krizhevsky2012imagenet,simonyan2014very,he2016deep}, it is highly data-hungry and requires massive labeled training data.
In contrast, human-level intelligence can achieve fast learning after observing only one or few instances~\cite{lake2011one}.
To relieve this gap, researchers have devoted efforts to few-shot learning problem, such as similarity metric~\cite{koch2015siamese,snell2017prototypical}, meta learning~\cite{finn2017model,rusu2019meta}, and augmentation~\cite{hariharan2017low,wang2018low,cheny2019multi}. 

\begin{figure*}[h]
	\centering
	\begin{tabular}{ccc}
		\includegraphics[width=0.23\textheight]{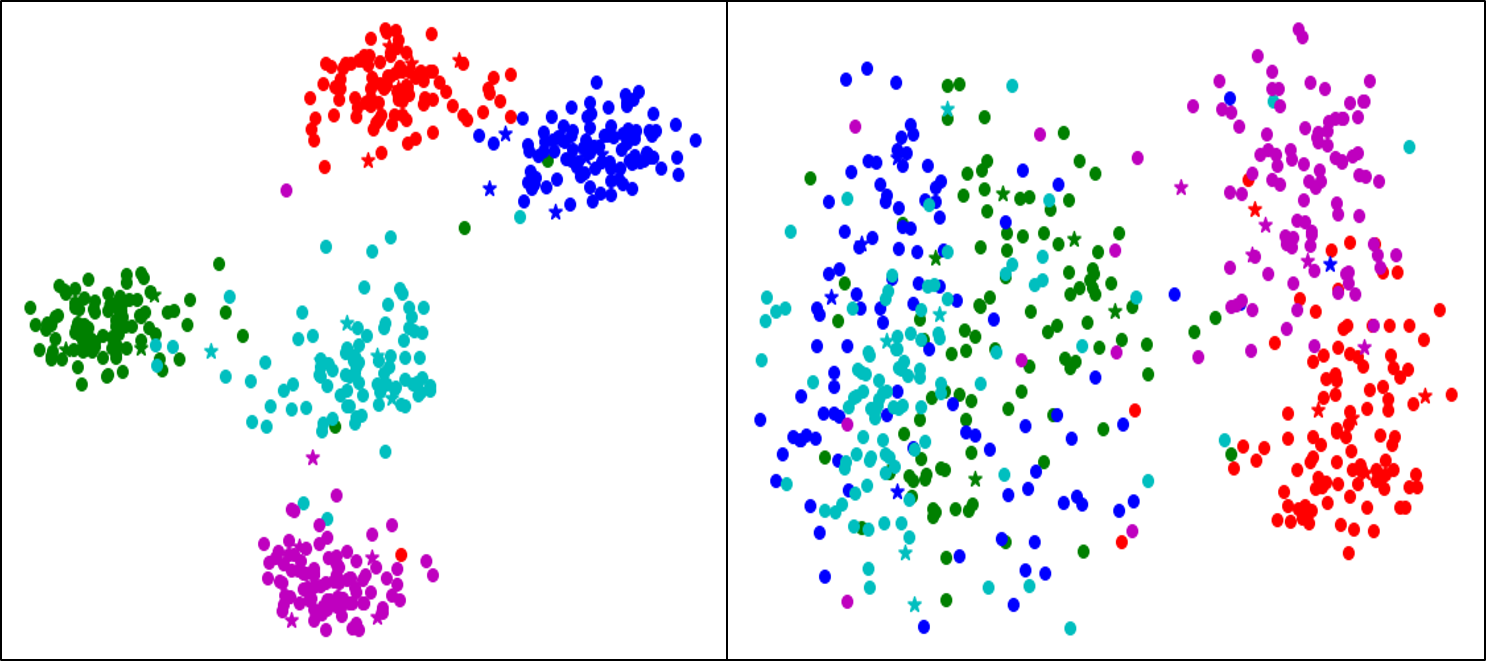} & 
		\includegraphics[width=0.23\textheight]{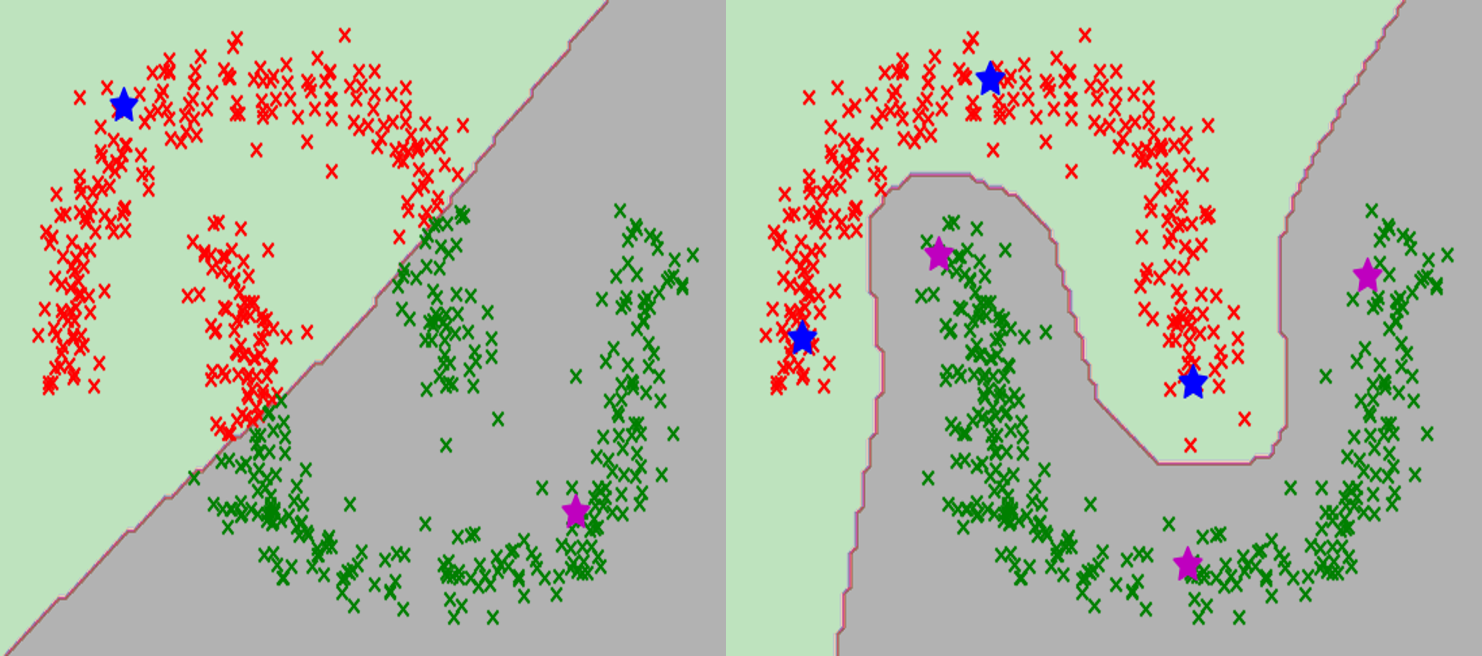} &
		\includegraphics[width=0.28\textheight]{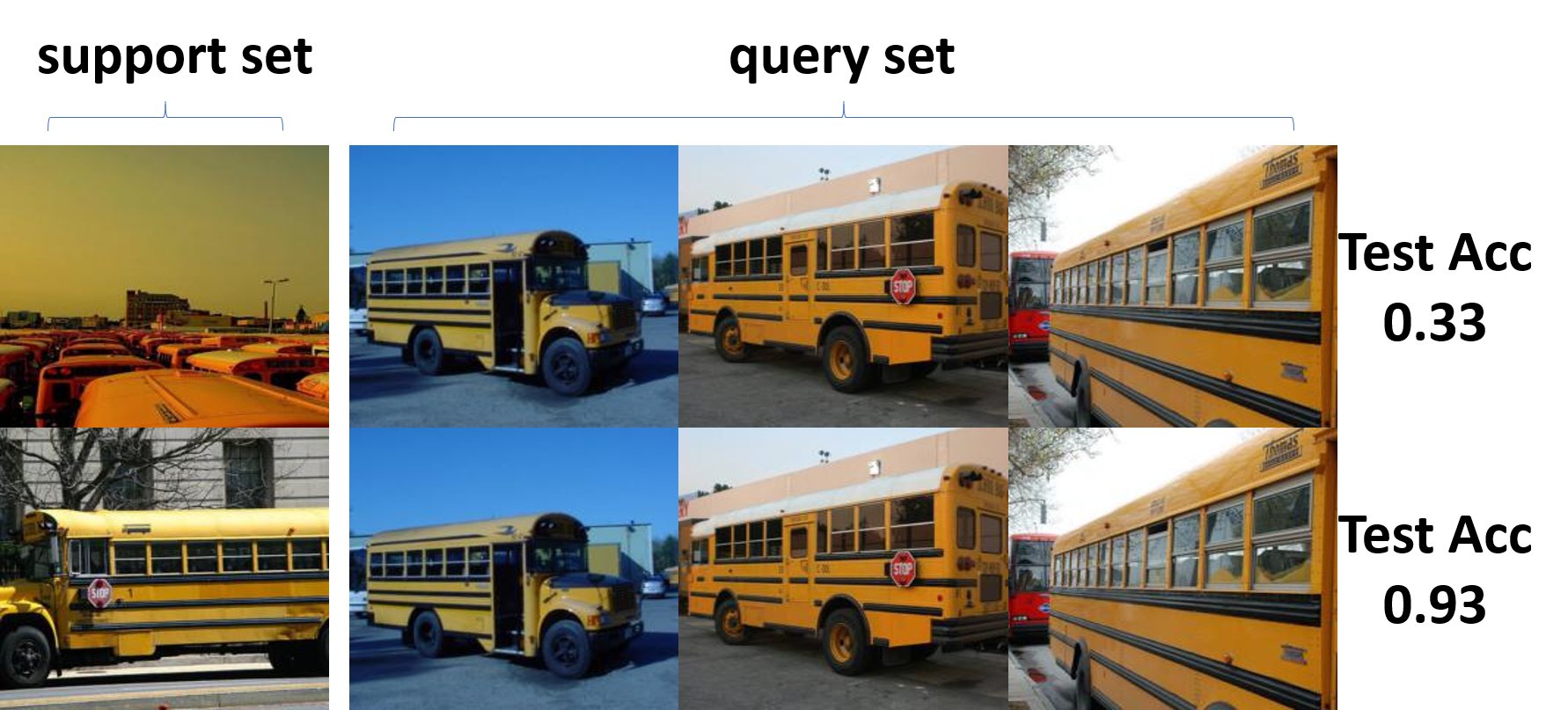}\\
		(a) Feature deterioration &
		(b) Data deficiency &
		(c) Representative training data
	\end{tabular}
	\caption{Challenges on few-shot image recognition. 
		(a) T-SNE visualization of ProtoNet's features on MiniImageNet. The feature discriminative property dramatically decreases from training (left) to novel testing (right) classes. 
		(b) Classifier decision boundary on the double moon toy dataset. It is difficult to learn a good decision boundary on a single labeled instance (marked with a star), but doable on more and representative labeled instances.
		(c) The performance comparison on different support sets. First row: the model performs badly on a random labeled data. Second row: the performance can be improved a lot if we use a representative training data. \label{Fig:challenges}}
\end{figure*}

However, the few-shot learning problem still remains challenging.
We argue that the main challenge comes from two aspects:
(1) \textbf{feature deterioration}. On the one hand, it's hard to learn a good feature representation based on a few of labeled data. On the other hand, the pre-trained feature representation on a dataset may lose its discriminative property on novel classes.
For example, as illustrated in Fig.~\ref{Fig:challenges} (a), the ProtoNet~\cite{snell2017prototypical} features deteriorate from training classes (left figure) to novel testing classes (right figure).
The model which works well on training classes may not have good performance on novel classes. 
(2) \textbf{Data deficiency}. A single or a few of labeled data could not represent the true data distribution. 
The biased distribution makes it difficult to learn a good decision boundary even with a decent feature representation.
This concept is showed in Fig.~\ref{Fig:challenges} (b).
Given only one labeled data (marked with stars) in each class (left figure), it's difficult to learn an accurate decision function. 
With more representative labeled data (right figure), however, the decision function can get improved.
Most of previous approaches treat their model as a black box and actually suffer from both issues. 
A recent work~\cite{chen2019closerfewshot} shows that existing approaches like ProtoNet~\cite{snell2017prototypical} and MAML~\cite{finn2017model} could be even beaten by the standard transfer learning baseline in some scenarios. 

In this paper, we focus on the data deficiency issue and use a sophisticated network architecture (e.g. ResNet) to alleviate feature deterioration as suggested by Kornblith et al.~\cite{kornblith2019better}.
We come up with our approach from an important observation.
As illustrated in Fig.~\ref{Fig:challenges} (c), the only difference between two few-shot episodic testing (i.e. first row and second row) is the labeled images (i.e. support set), but the testing accuracy varies  significantly.
It seems that representative training data play a vital role in few-shot learning.
Indeed, it is hard to precisely define what a representative sample is since we may not have prior knowledge about the novel classes in practice.  
However, it should be generally helpful if model has the chance to see more data. 

Based on this observation, we propose a continual local replacement approach which leverages the content of unlabeled images to constantly alter the labeled images. 
In particular, some regions of the labeled images will be replaced by unlabeled images and its contents can be learned through a supervised way.
Note that the labeled and unlabeled images may be semantically different (i.e. different classes). 
In this case, the replacement can be seen as artifacts or random erasing~\cite{zhong2020random} which brings limited benefits. 
To select informative unlabeled data, we borrow the idea of pseudo-labeling~\cite{lee2013pseudo} and design an algorithm which makes the data selection on the fly before each training epoch.
This brings two advantages.
First, semantically similar data will be selected to enhance labeled images, and thus the model can learn new semantic information to improve itself. 
Second, different data will be continually selected which diversifies semantic information and the supervised signals in the embedding space could be enlarged.

In this paper, the main technical contribution is a continual local replacement strategy which effectively addresses the data deficiency issue and learns a better classifier decision boundary. 
Our algorithm is built upon standard transfer learning.
It is simple, scalable and interpretable. 
It also can be seen as an extension of the baseline method in~\cite{chen2019closerfewshot}.
Extensive experiments show that this simple yet effective approach can significantly improve the baseline's absolute accuracy by $8\% \sim 15\% $ and achieve state-of-the-art results on various image recognition datasets.
Source code has been made available in \url{https://github.com/Lecanyu/ContinualLocalReplacement}.  
We hope this approach can be a strong baseline for future research.

\section{Related Work}
Our method takes unlabeled images to alter labeled ones for few-shot image recognition.
Such a strategy is closely related to few-shot image recognition, semi-supervised learning, and data augmentation.  
We briefly review the most relevant works below.

\textbf{Few-shot image recognition}.
The goal of few-shot image recognition is to endow models with the ability to recognize novel classes and datasets where only a limited amount of labeled data is available. 
Many approaches have been proposed for this goal. 
The former work like~\cite{fei2006one,salakhutdinov2012one} presented the Bayesian model for few-shot image recognition.
The recent metric learning approaches~\cite{koch2015siamese,snell2017prototypical,sung2018learning} learn metrics from pairwise comparisons between image instances.
And the meta learning methods~\cite{finn2017model,zhou2018deep,rusu2019meta} learn a good model initialization for the few-shot adaptation. 
Besides, the attention-based~\cite{wang2017multi}, graph-based~\cite{garcia2018few,liu2019learning}, and memory-based~\cite{santoro2016one,cai2018memory} strategies have also been investigated.

\textbf{Semi-supervised learning}.
Unlike standard supervised learning, it attempts to leverage unlabeled data to help learning, especially when labeled data are insufficient. 
The previous work~\cite{grandvalet2005semi} proposed the entropy regularization to encourage learning low-density separations between classes.
The pseudo-label approach~\cite{lee2013pseudo} iteratively trains and updates the model on unlabeled data with guessed labels.
The perturbation and consistency regularization strategies~\cite{sajjadi2016regularization,miyato2018virtual} perturb unlabeled data and force consistent output to exploit potential data distribution.
A hybrid approach~\cite{berthelot2019mixmatch} integrates previous strategies and achieves the new state-of-the-art.
Although these semi-supervised methods are usually inapplicable on few-shot learning where the number of labeled data is extremely small, they recently draw researchers' attention to consider taking advantage of unlabeled data for few-shot learning.
For example, Ren et al.~\cite{ren2018meta} proposes a variant of ProtoNet which uses unlabeled data to further adjust the learned prototypes.
Gidaris et al.~\cite{gidaris2019boosting} applies some self-supervision methods like rotation to boost few-shot learning.  
And Dhillon et al.~\cite{dhillon2020baseline} takes entropy regularization method to improve few-shot classification performance.
In this paper, we take the idea of pseudo-labeling with image local replacement to address the data deficiency issue in few-shot learning.

\textbf{Data augmentation}.
Data augmentation is widely adopted in various machine learning problems.
In image processing, the typical and standard augmentations are rotating, flipping, cropping, color jittering and etc.
Such augmentations can endow models with better generalization~\cite{krizhevsky2012imagenet}.
More sophisticated augmentation methods have also been explored, such as image synthesis~\cite{wang2018low}, random erasing~\cite{zhong2020random}, feature hallucination and augmentation~\cite{hariharan2017low,cheny2019multi}. 

In these existing methods, the most relevant works are~\cite{chen2019image,chen2019image_mixup}. 
They introduced various patch-level image augmentation methods like mixup and replacement.
However, they ignored the diversity of augmented images and used a fixed augmented set even though a learning-based augmentation strategy is adopted in~\cite{chen2019image_mixup}.
By contrast, we apply image local replacement on the fly in each training epoch.
The semantic information could be directly diversified through this way.


\section{Continual Local Replacement}
\label{Sec:CLR}
\subsection{Preliminary}
The few-shot image recognition problem can be described on training and testing datasets.
The training dataset has abundant labeled classes $C_{train}$.
After training on samples with labels from $C_{train}$, the goal is to produce a model for recognizing a disjoint set of novel classes $C_{test}$ (i.e. $C_{train} \cap C_{test} = \emptyset$) for which only a single or few of labeled samples are available.
Formally, let $(x, y)$ denote an image and its label respectively. 
The training dataset $D_{train}=\{(x_i, y_i)\}_{i=1}^{N_{train}}$, where $y_i \in C_{train}$.
The testing dataset $D_{test}=\{(x_i, y_i)\}_{i=1}^{N_{test}}$, where $y_i \in C_{test}$.
Most of recent works on few-shot learning follow the episodic paradigm~\cite{vinyals2016matching}.
In particular, the episodic testing consists of hundreds of independent testing episodes.
For each episode, it will randomly sample an episodic testing dataset $d_{test}=\{(x_i, y_i)\}_{i=1}^{m} \subset D_{test}$ which contains $n$ random classes from $C_{test}$ and total $m=(k+t)\times n$ images where $k$ and $t$ are the number of labeled images and testing images per class respectively. 
In few-shot literature, the $n\times k$ labeled images and $n \times t$ testing images are also referred to as the \emph{support} set $d_{support}$ and \emph{query} set $d_{query}$, respectively. 


\subsection{Image Local Replacement}
Our primary goal is to provide a chance for the model to see more data.
We fulfill it via introducing the content of unlabeled images for learning.
The image local replacement is a robust way to introduce new semantic information.
\begin{figure}[h]
	\centering
	\begin{tabular}{ccc}
		\includegraphics[width=0.1\textheight]{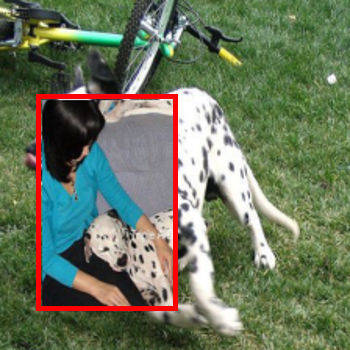} &
		\includegraphics[width=0.1\textheight]{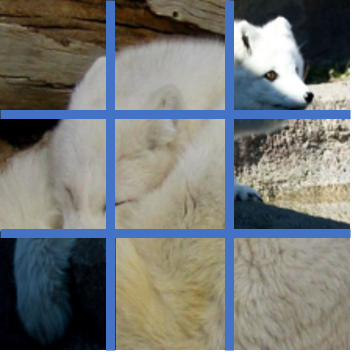} &
		\includegraphics[width=0.1\textheight]{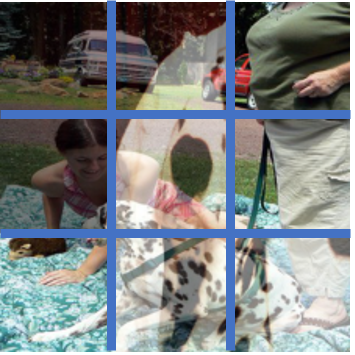}\\
		(a) RandEra & (b) BlockAug  & (c) BlockDef  
	\end{tabular}
	\caption{Various image local replacement methods. 
		(a) Random erasing~\protect\cite{zhong2020random}. The red box area comes from another semantic relevant image.
		(b) Block augmentation~\protect\cite{chen2019image}. The image is divided into 9 blocks and some of them are replaced by other images.
		(c) Block deformation~\protect\cite{chen2019image_mixup}. Some sub-blocks are linearly mixed with other images. \label{Fig:augmentations}}
\end{figure}
There are already some replacement methods. 
For example, a random region of an image is replaced by another one as showed in Fig.~\ref{Fig:augmentations}(a).
The original image can be divided to several blocks and some of them are either replaced like Fig.~\ref{Fig:augmentations}(b) or linearly mixed with other images like Fig.~\ref{Fig:augmentations}(c).
All these replacement methods are applicable for our purpose.
Note that the random \emph{local} replacement is crucial for the robustness.
When two images are semantically different, the replaced regions may be interpreted as partial occlusions.
This can be seen as an augmentation to improve the generalization ability~\cite{chen2019image,zhong2020random}.  
When two images are semantically similar, the model has a chance to learn new semantic information from replaced regions.
Except the existing replacement methods, we may also design other fancy methods as long as it can satisfy these two objectives. 
But this is beyond the scope of this paper since our technical contribution mainly lies in the following \emph{continual} replacement approach. 

\subsection{Training}
\begin{figure}[h]
	\centering
	\begin{tabular}{c}
		\includegraphics[width=0.35\textheight]{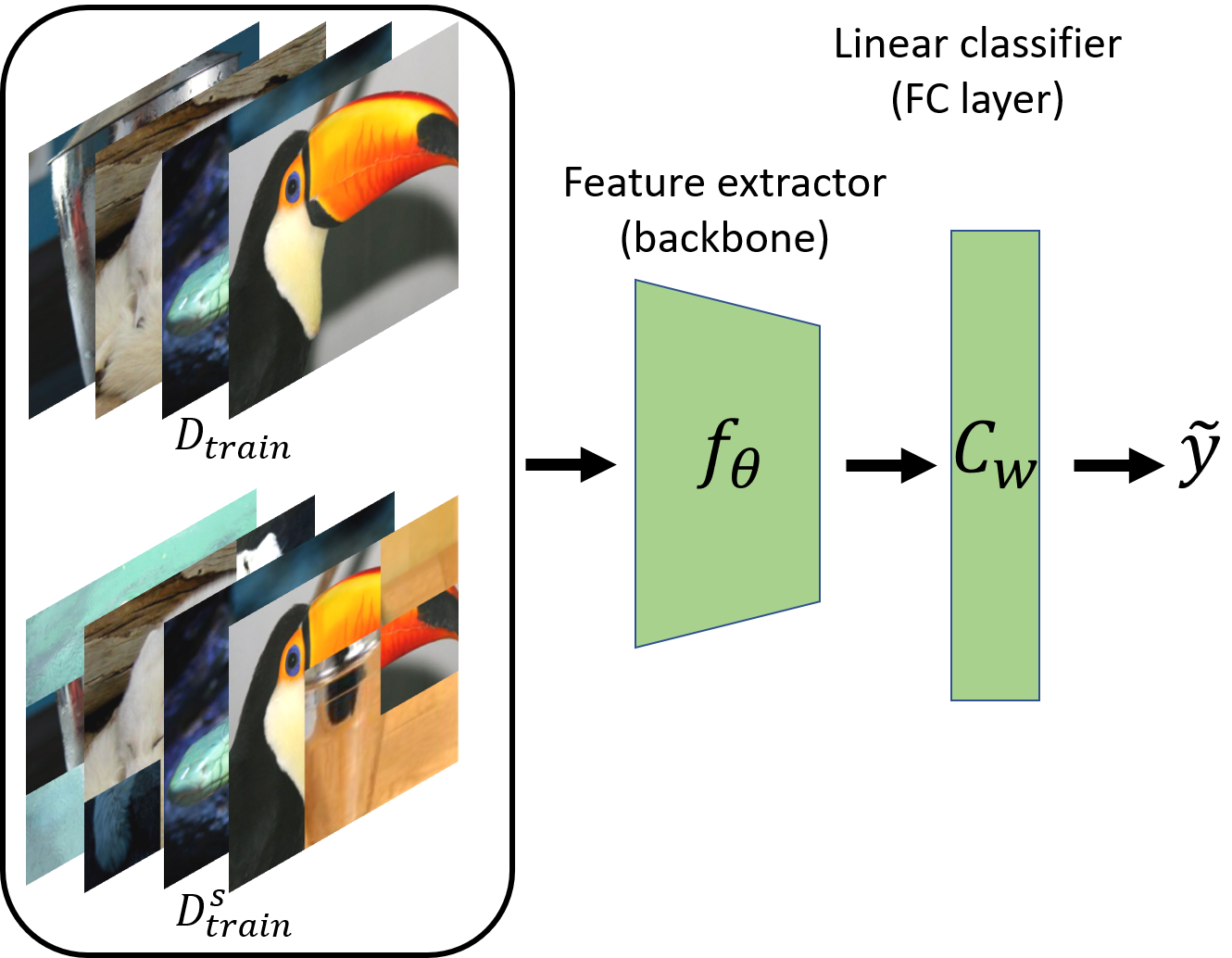}
	\end{tabular}
	\caption{Training stage. Original and locally replaced images are fed into CNN simultaneously. \label{Fig:training}}
\end{figure}
In the training stage, we first apply local replacement on the original training set $D_{train}$ to synthesize an augmented version training set $D_{train}^s$. 
Specifically, for each image $x_i \in D_{train}$, we randomly select another image $x_j \in D_{train}, i \ne j$ with $0.5$ probability that $y_i=y_j$ (i.e. two different images with the same label). 
Then, one or several local regions of $x_i$ will be replaced by $x_j$ to synthesize a new image $x_i^s$. 
The new image $x_i^s$ is still with the original label $y_i$.
The reason why we introduce $D_{train}^s$ in training is that we want the model to understand such local replacement and make it robust to partial occlusions.

After building up $D_{train}^s$, both the original images and synthesized images are simultaneously fed into CNN for training.
This procedure is illustrated in Fig.~\ref{Fig:training}.
Specifically, we optimize the loss function Eq. \ref{Eqn:train_loss} during the training:
\begin{equation}
\label{Eqn:train_loss}
\arg\min_{\theta, w} \sum_{D_{train}, D_{train}^s} \ell(C_w(f_{\theta}(x_i)), y_i),
\end{equation} 
where $\ell(\cdot,\cdot)$ is the standard cross-entropy loss.
$f_{\theta}$ denotes the feature extractor backbone with trainable parameters $\theta$ and $C_{w}$ is a classifier (e.g. a fully connected layer) with parameters $w$.

\subsection{Fine Tuning}
\begin{figure}[h]
	\centering
	\begin{tabular}{c}
		\includegraphics[width=0.35\textheight]{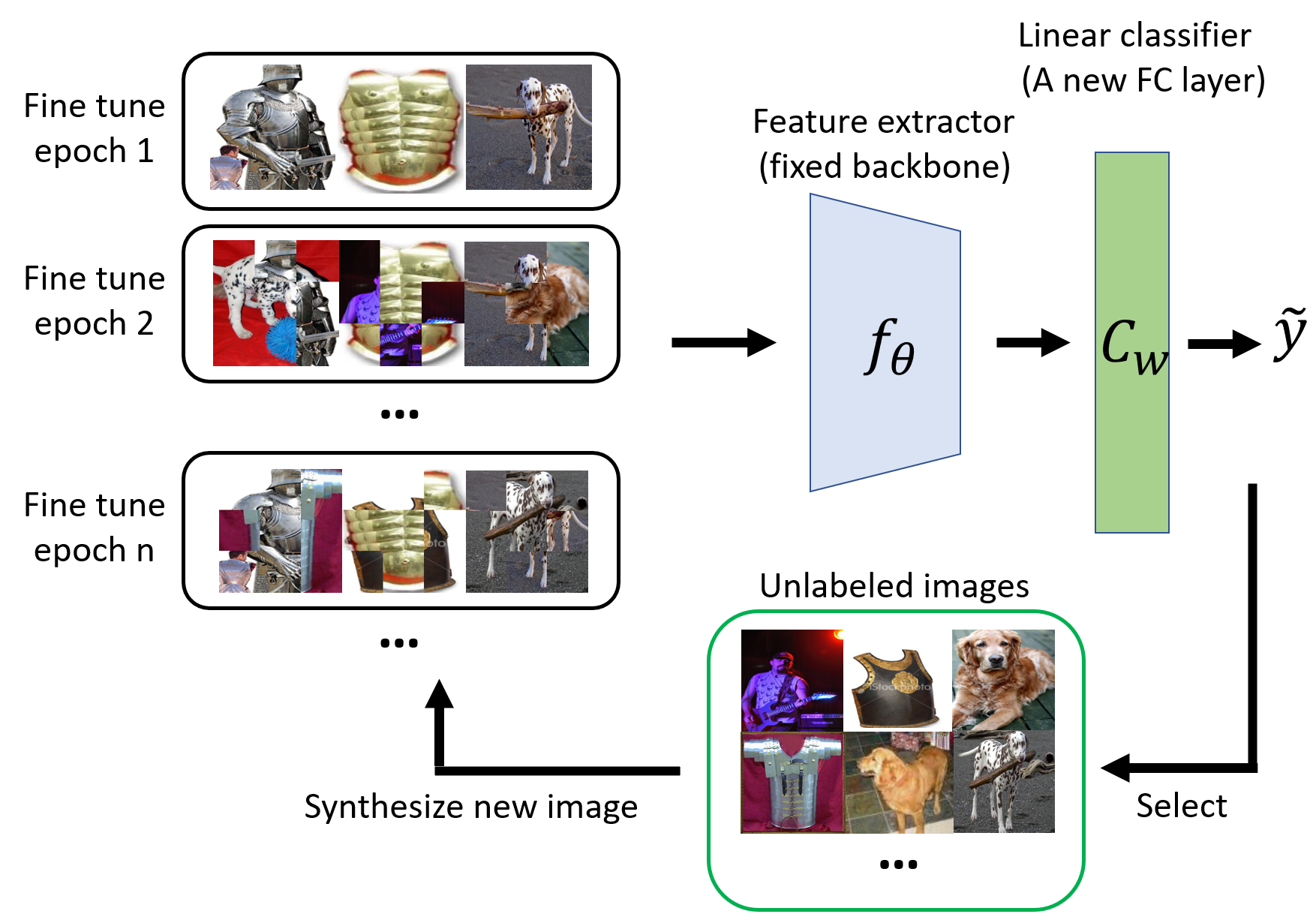}
	\end{tabular}
	\caption{Fine-tune stage. Before each fine-tune epoch, the current model guesses labels on unlabeled images and selects semantically similar instances for the local replacement and new image synthesis. \label{Fig:fine_tune}}
\end{figure}
After training, we fine tune the model to recognize novel classes in $D_{test}$.
For each testing episode, additional $u$ images per class will be randomly sampled as the unlabeled set $d_{unlabel}=\{x_i\}_{i=1}^{n\times u}$.
In other words, there will be total $m=(k+t+u)\times n$ images including support and query sets.

We follow the standard transfer learning procedure to fine tune the trained model on $D_{train}$ and $D_{train}^s$.
Specifically, suppose the pre-trained feature extractor is reasonably good.
We fix the pre-trained feature extractor parameters $\theta$ and create a new classifier with random initialized parameters $w$.  
The image local replacement is constantly applied when tuning the classifier.
At the first fine-tune epoch, the original support set is fed into the model and classifier parameters $w$ are updated by minimizing the cross-entropy loss.
Before the following tuning epochs, the latest model makes predictions for all unlabeled images.
The pseudo labels $\widetilde{y}_i$ are assigned to unlabeled set $d_{u}=\{(x_i, \widetilde{y}_i)\}$. 
Then, local regions of an image $x_i \in d_{support}$ will be replaced by an unlabeled image $x_j \in d_{unlabel}$ with $\widetilde{y}_j=y_i$ to synthesize a new image $x_i^s$.
And an augmented support dataset $d_{support}^s=\{(x_i^s, y_i)\}$ can be obtained. 
The $d_{support}^s$ is fed into CNN to fine tune the classifier in the next epoch and Eq. \ref{Eqn:fine_tune_loss} will be optimized:
\begin{equation}
\label{Eqn:fine_tune_loss}
\arg\min_{w} \sum_{d_{support}^s} \ell(C_w(f_{\theta}(x_i)), y_i).
\end{equation} 
The complete fine tune algorithm is summarized in Algorithm.~\ref{Alg:fine_tune}.
\begin{algorithm}  
	\caption{The fine tune algorithm}  
	\label{Alg:fine_tune}  
	\begin{algorithmic}  
		\REQUIRE{The pretrained feature extractor $f_\theta$; support, unlabeled sets $d_{support}, d_{unlabel}$}
		\ENSURE{The linear classifier $C_w$}
		\STATE Randomly initialize a new linear classifier $C_w$.
		\STATE Initialize $d_{support}^s \gets d_{support}$.
		\FOR{ epoch$=1, 2, ...$}
		\FOR{ mini-batch in $d_{support}^s$ } 
		\item Optimize classifier $C_w$ using Eq.~\ref{Eqn:fine_tune_loss}.
		\ENDFOR
		\item Predict labels $\widetilde{y}_i$ for each image $x_i \in d_{unlabel}$. 
		\item Select a subset $d_{unlabel}^{'} \subset d_{unlabel}$ which is semantically similar with $d_{support}$.  
		\item $d_{support}^s \gets$ locally replace $d_{support}$ using $d_{unlabel}^{'}$ 
		\ENDFOR
	\end{algorithmic}  
\end{algorithm}

This approach is robust to wrong predictions on unlabeled images (i.e. $\widetilde{y}_i \ne y_i$) by controlling the area of local replacement.
We set a threshold $\alpha$ so that the maximum $\alpha\%$ area of the original image is allowed to be locally replaced.
When using a small $\alpha$, the wrong prediction doesn't hurt performance because it can be interpreted as partial occlusions or artifacts~\cite{chen2019image,zhong2020random}.
However, small replaced regions introduce little semantic information.  
So we prefer to set a larger $\alpha$ so that the model can learn new semantic contents. 
Indeed, large replaced regions may lead to a risk of learning wrong contents and the fine-tune procedure may be trapped in poor local minima.
We alleviate this issue by randomly determining the position and the size of replaced regions.
Note that the fine-tune procedure can still converge quickly since $d_{support}^s$ always contain the contents of the original support set.
Empirically, this simple strategy works well and $\alpha=65\%$ can roughly achieve the best trade-off across all experiment benchmarks.

\begin{figure}[h]
	\centering
	\begin{tabular}{c}
		\includegraphics[width=0.38\textheight]{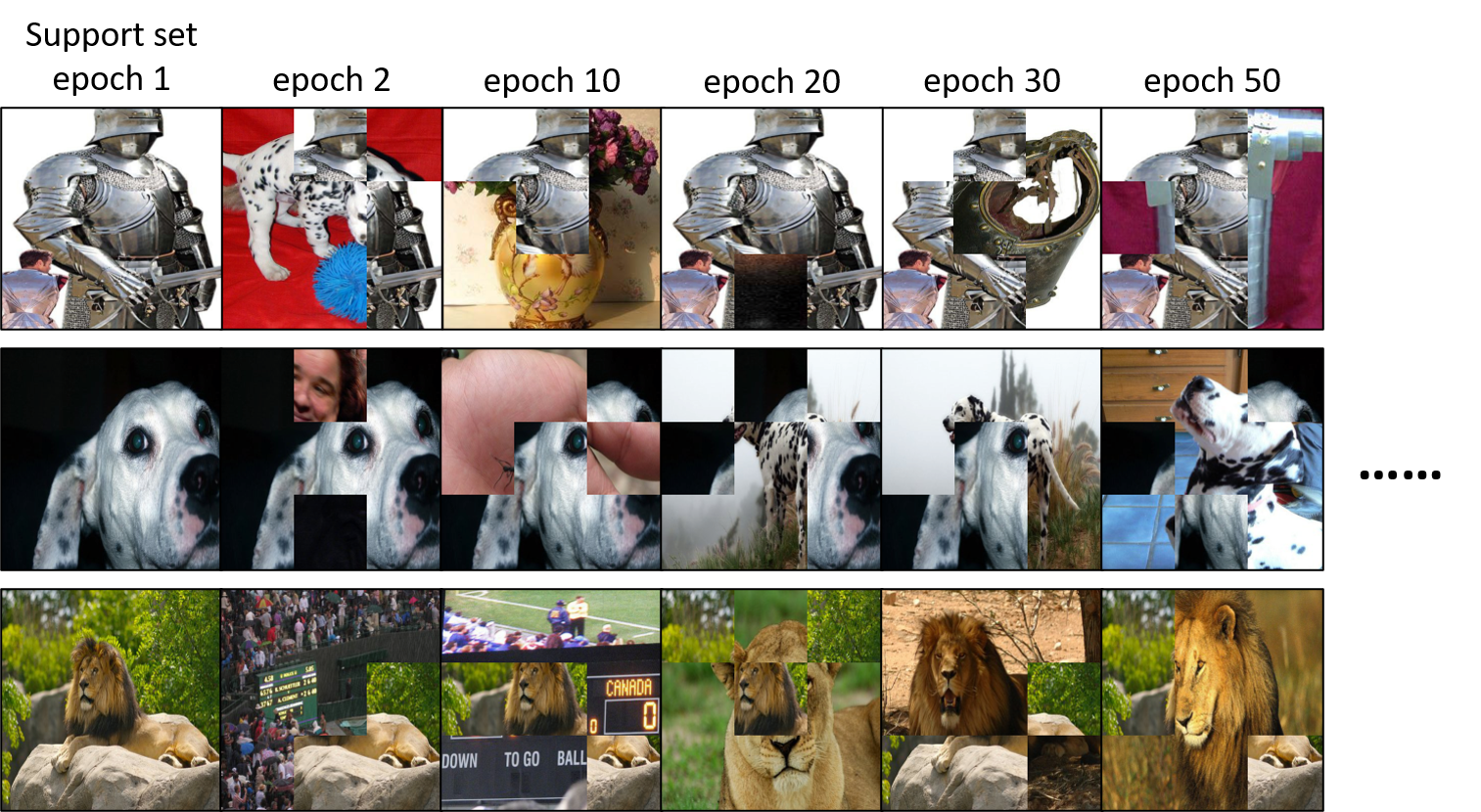}
	\end{tabular}
	\caption{Continual local replacement process in MiniImagenet 1-shot setting. On fine-tune stage, CLR constantly select semantically similar images to locally replace images in support set. The selection could be wrong at the beginning, but it gets improved with more fine tune iteration steps. Correct selection further helps model learning on following iterations. \label{Fig:CLR_process}}
\end{figure}

Fig. \ref{Fig:CLR_process} illustrates the continual local replacement process in MiniImagenet 1-shot testing.
The first column is the original support set. 
With the fine-tune iterating, model can select semantically similar images more accurately which introduces new information to further help learning and boost its performance.  

\section{Experiments}
The experiments are conducted on four widely used few-shot learning benchmarks: MiniImageNet~\cite{ravi2017optimization}, TieredImageNet~\cite{ren2018meta}, Caltech-256~\cite{griffin2007caltech} and CUB-200~\cite{wah2011caltech}.
These datasets cover from general objects (e.g. MiniImageNet) to fine grained bird species (e.g. CUB-200) and they can provide a comprehensive evaluation for our method.
We implement our approach based on a recent testbed~\cite{chen2019closerfewshot} and follow its basic experimental settings, dataset split and hyper-parameters without any other tricks or hyper-parameter tuning.
We use ResNet-18 backbone with a single fully connected classifier (FC layer) on all datasets.
The ResNet-18 backbone is only trained on the training set $D_{train}$ and the augmented version $D_{train}^s$ from scratch. 
Note that $D_{train}$ and $D_{train}^s$ are essentially the same dataset and thus the total capacity of training information is not changed. 
During the episodic testing stage, the parameters of backbone are fixed and a new FC layer will be fine tuned.
We adopt the image block augmentation~\cite{chen2019image} as the local replacement strategy. But we also evaluate other local replacement methods like random erasing \cite{zhong2020random} and block deformation \cite{chen2019image_mixup} in the following ablation study.
Additional $u=15$ unlabeled images are randomly sampled and maximum 4 and 6 random blocks are allowed to be replaced during the training and fine-tune stages, respectively.

\subsection{Standard Few-shot Learning Evaluation}
We follow the episodic testing protocol~\cite{vinyals2016matching} for the standard few-shot learning evaluation.

\begin{table*}[h]
	\centering
	\scriptsize
	\setlength\tabcolsep{4.0pt}
	\caption{1-shot 5-way and 5-shot 5-way testing results on MiniImageNet, TieredImageNet, CUB-200, Caltech-256 datasets. Most of results come from original paper. 
		The results of methods marked with '$\dagger$' come from a recent testbed \protect\cite{chen2019closerfewshot} or obtained by using their released code.   \label{Tab:results}}
	\begin{tabular}{|c|c|cc|cc|cc|cc|}
		\hline
		\multirow{2}{*}{Methods} & \multirow{2}{*}{Arch.} & \multicolumn{2}{c|}{ MiniImageNet (\%) } & \multicolumn{2}{c|}{ TieredImageNet (\%) } & \multicolumn{2}{c|}{ CUB-200 (\%) } & \multicolumn{2}{c|}{ Caltech-256 (\%) } \\
		& & 1-shot & 5-shot & 1-shot & 5-shot & 1-shot & 5-shot & 1-shot & 5-shot  \\
		\hline 
		S.S. ProtoNet~\cite{ren2018meta} & CONV4 & 50.41 $\pm$ 0.31 & 64.39 $\pm$ 0.24 & 52.39 $\pm$ 0.44  & 69.88 $\pm$ 0.20 &	 -  &  -  & -  &  -    \\
		\hline 
		TPN~\cite{liu2019learning} & CONV4 & 54.72 $\pm$ 0.84 & 69.25 $\pm$ 0.67 & 	59.91 $\pm$ 0.94  & 73.30 $\pm$ 0.75 &  -   & - & - & -  \\
		\hline 
		MAML$\dagger$~\cite{finn2017model} & ResNet-18 & 49.61$\pm$0.92  & 65.72$\pm$0.77 & 	50.80 $\pm$ 1.00 & 72.69 $\pm$ 0.86  &	 69.96 $\pm$ 1.01  & 82.70 $\pm$ 0.65 & 57.33 $\pm$ 1.00  & 75.77 $\pm$ 0.70   \\
		\hline 
		MatchNet$\dagger$~\cite{vinyals2016matching} & ResNet-18 & 52.91 $\pm$ 0.88  & 68.88 $\pm$ 0.69 &  53.18 $\pm$ 0.90 & 72.77 $\pm$ 0.71 &	 72.36 $\pm$ 0.90  & 83.64 $\pm$ 0.60 & 62.24 $\pm$ 0.89 & 77.92 $\pm$ 0.66  \\
		\hline 
		ProtoNet$\dagger$~\cite{snell2017prototypical} & ResNet-18 & 54.16 $\pm$ 0.82  & 73.68 $\pm$ 0.65 & 	52.55 $\pm$ 0.88  & 75.35 $\pm$ 0.74 &	71.88 $\pm$ 0.91  & \textbf{87.42 $\pm$ 0.48} & 60.17 $\pm$ 0.90 & 80.56 $\pm$ 0.64 	\\
		\hline 
		RelationNet$\dagger$~\cite{sung2018learning} & ResNet-18	& 52.48$\pm$0.86 & 69.83$\pm$0.68  & 47.01 $\pm$ 0.91 & 69.29 $\pm$ 0.81 &	67.59 $\pm$ 1.02 & 82.75 $\pm$ 0.58 & 55.72 $\pm$ 0.90 & 77.42 $\pm$ 0.68  \\
		\hline 
		Baseline$\dagger$~\cite{chen2019closerfewshot} & ResNet-18 & 51.75 $\pm$ 0.80 & 74.27 $\pm$ 0.63 &  65.12 $\pm$ 0.86  & 85.05 $\pm$ 0.63  & 65.51 $\pm$ 0.87 & 82.85 $\pm$ 0.55 & 57.72 $\pm$ 0.88 & 79.06 $\pm$ 0.67 \\
		\hline 
		Baseline++$\dagger$~\cite{chen2019closerfewshot} & ResNet-18	& 51.87 $\pm$ 0.77 & 75.68 $\pm$ 0.63 & 64.83 $\pm$ 0.92 & 82.77 $\pm$ 0.72 & 67.02 $\pm$ 0.90 & 83.58 $\pm$ 0.54 & 56.72 $\pm$ 0.90 & 77.24 $\pm$ 0.67  \\
		\hline 
		BlockAug~\cite{chen2019image} & ResNet-18 & 58.80 $\pm$ 1.36 & 76.71 $\pm$ 0.72 & - & - & - & - & - & - \\
		\hline
		IDeMe-Net~\cite{chen2019image_mixup} & ResNet-18 & 59.14 $\pm$ 0.86 & 74.63 $\pm$ 0.74 &  -  & -  &  -  &  -  &  -  &  -  \\
		\hline 
		DEML~\cite{zhou2018deep} & ResNet-50 & 58.49 $\pm$ 0.91 & 71.28 $\pm$ 0.69  & - & - & 66.95 $\pm$ 1.06 & 77.11 $\pm$ 0.78 & 62.25 $\pm$ 1.00  & 79.52 $\pm$ 0.63  \\
		\hline 
		DualTriNet~\cite{cheny2019multi} & ResNet-18 & 58.12 $\pm$ 1.37 & 76.92 $\pm$ 0.69 &  -  &  -  &  69.61 $\pm$ 0.46 & 84.10 $\pm$ 0.35 & 63.77 $\pm$ 0.62  & 80.53 $\pm$ 0.46 \\
		\hline
		LEO~\cite{rusu2019meta} & WRN28-10  & 61.76 $\pm$ 0.08 & 77.59 $\pm$ 0.12 & 66.33 $\pm$ 0.05 & 81.44 $\pm$ 0.09  &		- & - & - & -  \\
		\hline 
		MetaOptNet~\cite{lee2019meta} & ResNet-12 & 61.41 $\pm$ 0.61 & 77.88 $\pm$ 0.46 & 	 65.36 $\pm$ 0.71 &  81.34 $\pm$ 0.52 &	- & - & - & -  \\ 
		\hline
		MTL~\cite{sun2019meta} & ResNet-12 & 61.2 $\pm$ 1.8 & 77.88 $\pm$ 0.46  & 65.60 $\pm$ 1.80 &  80.60 $\pm$ 0.90 &  -  &  -   &  -  &   -   \\ 
		\hline
		BoostFSL~\cite{gidaris2019boosting} & WRN28-10  & 63.77 $\pm$ 0.45 & 80.70 $\pm$ 0.33  & - & -  &  -  &  -   &  -  &   -  \\ 
		\hline
		BaseFSL~\cite{dhillon2020baseline} & WRN28-10  & 65.73 $\pm$ 0.68 & 78.40 $\pm$ 0.52  & 73.34 $\pm$ 0.71 &  85.50 $\pm$ 0.50  &  -  &  -   &  -  &   -  \\ 
		\hline 
		\hline
		Vanilla (ours) & ResNet-18 & 56.44 $\pm$ 0.81 & 78.18 $\pm$ 0.60  & 63.93 $\pm$ 0.89 & 84.66 $\pm$ 0.62 & 65.91 $\pm$ 0.88 & 83.45 $\pm$ 0.51 & 59.32 $\pm$ 0.88 & 79.95 $\pm$ 0.67 \\
		CLR+Imprinting (ours) & ResNet-18 & 63.54 $\pm$ 0.85 & 78.26 $\pm$ 0.61 & 73.29 $\pm$ 0.97  & 85.04 $\pm$ 0.64 &  \textbf{73.80 $\pm$ 0.92 } & \textbf{87.46 $\pm$ 0.48} &  \textbf{65.88 $\pm$ 0.97} & 81.16 $\pm$ 0.64  \\
		CLR (ours) & ResNet-18 & \textbf{66.33 $\pm$ 0.93} & \textbf{81.12 $\pm$ 0.60} & \textbf{74.76 $\pm$ 0.97} & \textbf{86.78 $\pm$ 0.66} & 68.00 $\pm$ 0.92 & 84.26 $\pm$ 0.53 & \textbf{65.74 $\pm$ 0.92} & \textbf{83.05 $\pm$ 0.60} \\
		\hline
	\end{tabular}
\end{table*}

\textbf{Baselines and competitors}.
There are a lot of existing few-shot learning methods.
We compare our approach with the most relevant ones.
The competitors include popular baselines like ProtoNet~\cite{snell2017prototypical}, MAML~\cite{finn2017model}, semi-supervised approaches like S.S. ProtoNet~\cite{ren2018meta}, BoostFSL~\cite{gidaris2019boosting}, augmentation-based methods like BlockAug~\cite{chen2019image}, IDeMeNet~\cite{chen2019image_mixup}, DualTriNet~\cite{cheny2019multi} and state-of-the-art approaches like BaseFSL~\cite{dhillon2020baseline}.
Since different backbone architectures could significantly affect the results, we report the used architecture for each method along with the 5-way 1-shot and 5-way 5-shot accuracy.

In addition to our Continual Local Replacement (CLR) approach, we also report the results of two variants: Vanilla and CLR+Imprinting.
The vanilla version is a simple transfer learning approach. 
The model is trained on $D_{train}$ and $D_{train}^s$, but it is fine tuned on the original support set $d_{support}$ without any local replacement. 
So it can be seen as a baseline of our approach.
Our algorithm has good scalability and can be easily combined with existing methods such as weight imprinting~\cite{qi2018low}.
The imprinting version is inspired from a recent weight initialization method~\cite{qi2018low}.
We introduce the imprinting technique to our framework since it learns similarity metric and can be seen as a complement for linear classifier.
Specifically, the imprinting method normalizes the embedded features and the parameters of FC layer during the training.
When the features and parameters are normalized, 
the learning objective is equivalent to maximizing cosine similarity.
Then, on the fine tuning stage, it takes the average of features on the support set as the initial parameters of the new FC layer.

\textbf{Results}.
Comparison results on all datasets are shown in Table~\ref{Tab:results}. 
We can observe that: 
(1) our proposed approach can achieve the best performance on all datasets.
our method outperforms two most related methods BlockAug~\cite{chen2019image} and IDeMe-Net~\cite{chen2019image_mixup} with a clear improvement ($\approx 7\%$).
This validates the effectiveness of our continual local replacement strategy. 
(2) Our simple vanilla variant shows competitive performance. 
For example, the vanilla method has comparable or better performance over existing methods on MiniImageNet, TieredImageNet and Caltech-256 in 5-way 5-shot testing. 
This indicates that the deeper backbone can learn better feature representation in transfer learning and the feature deterioration issue can be alleviated by applying a sophisticated architecture.
The similar conclusion is also observed in several recent works~\cite{kornblith2019better,chen2019closerfewshot}.
With continual local replacement (i.e. CLR or CLR+Imprinting), the performance can be further improved, especially in 1-shot setting.
This means that the CLR technique can constantly improve our baseline. 
(3) 
The imprinting variant is more effective on fine grained recognition tasks than general object recognition tasks.  
This variant demonstrates strong performance on CUB-200 dataset but is slightly worse than CLR on other three general object recognition datasets.
Interestingly, some previous metric-based learning methods like ProtoNet also show strong results on CUB-200.
Since Imprinting and ProtoNet explicitly maximize the cosine and Euclidean similarity respectively, 
we may think these metric-based learning strategies are suitable for fine grained classification tasks through reducing the intra-class variation~\cite{qi2018low,chen2019closerfewshot}. 

\subsection{Ablation Study}
We conduct ablation studies to help understand our approach better.

\textbf{The number of replaced blocks.}
We adopt the image block augmentation \cite{chen2019image} as the local replacement method.
The replaced area can be controlled by the number of replaced blocks.
We evaluate our approach with different numbers of replaced blocks on both training and fine tuning stages.
As shown in Table~\ref{Tab:diff_num_block}, the best results are achieved when the maximum numbers of replaced blocks on training/fine-tuning are roughly 3 and 6.
This indicates that it is necessary to replace some blocks to build up $D_{train}^s$ for training to obtain the best results on the fine tune stage. 
\begin{table}[h]
	\centering
	\scriptsize
	\caption{The performance of different number of replaced blocks on MiniImageNet 5-shot 5-way setting.  \label{Tab:diff_num_block}}
	\begin{tabular}{|c|c|c|c|c|c|}
		\hline
		& \multicolumn{5}{c|}{ Fine tuning } \\
		\hline
		\multirow{5}{*}{\rotatebox{90}{Training}} &  & 0 & 3 & 6 & 9 \\
		\cline{2-6}
		& 0 & 76.96 $\pm$ 0.60 & 78.11 $\pm$ 0.61  & 78.68 $\pm$ 0.61 & 79.16 $\pm$ 0.66 \\
		\cline{2-6} 
		& 3 & 77.16 $\pm$ 0.64  & 79.23 $\pm$ 0.60 & \textbf{80.83 $\pm$ 0.60} & 80.14 $\pm$ 0.62 \\
		\cline{2-6}
		& 6 & 76.44 $\pm$ 0.61 & 79.26 $\pm$ 0.61 & 80.07 $\pm$ 0.62  & 79.23 $\pm$ 0.64 \\
		\cline{2-6}
		& 9 & 76.41 $\pm$ 0.66  & 78.95 $\pm$ 0.56 & 79.15 $\pm$ 0.60 & 78.87 $\pm$ 0.68 \\
		\hline
	\end{tabular}
\end{table}

\textbf{The effect of different components.}
To verify the effectiveness of our CLR strategy, two different local replacement configurations: CLR RandEra, CLR BlockDef, and some variants of CLR: Vanilla, One Time Local Replacement (OTLR), CLR without Pseudo Labeling (CLR w/o PL), CLR with Ground Truth Label (CLR w GT), Continual All Replacement (CAR) are evaluated and compared.

Specifically, 
The CLR RandEra and CLR BlockDef use random erase \cite{zhong2020random} and block deformation \cite{chen2019image_mixup} as the local replacement method respectively which are introduced in Section \ref{Sec:CLR}.
The OTLR selects unlabeled images to replace some blocks in labeled images only once.
It can be seen as a direct augmentation method like BlockAug~\cite{chen2019image} without continual replacement.
The CLR w/o PL randomly selects unlabeled images to apply local replacement continually, whereas CLR w GT uses ground truth label to select unlabeled images which can be seen as an oracle and the upper bound of our method.
The CAR use the whole of unlabeled image to tune classifier instead of replacing local regions. 
This strategy can be seen as a straightforward pseudo labeling method.

\begin{figure}[h]
	\centering
	\begin{tabular}{c}
		\includegraphics[width=0.45\textwidth]{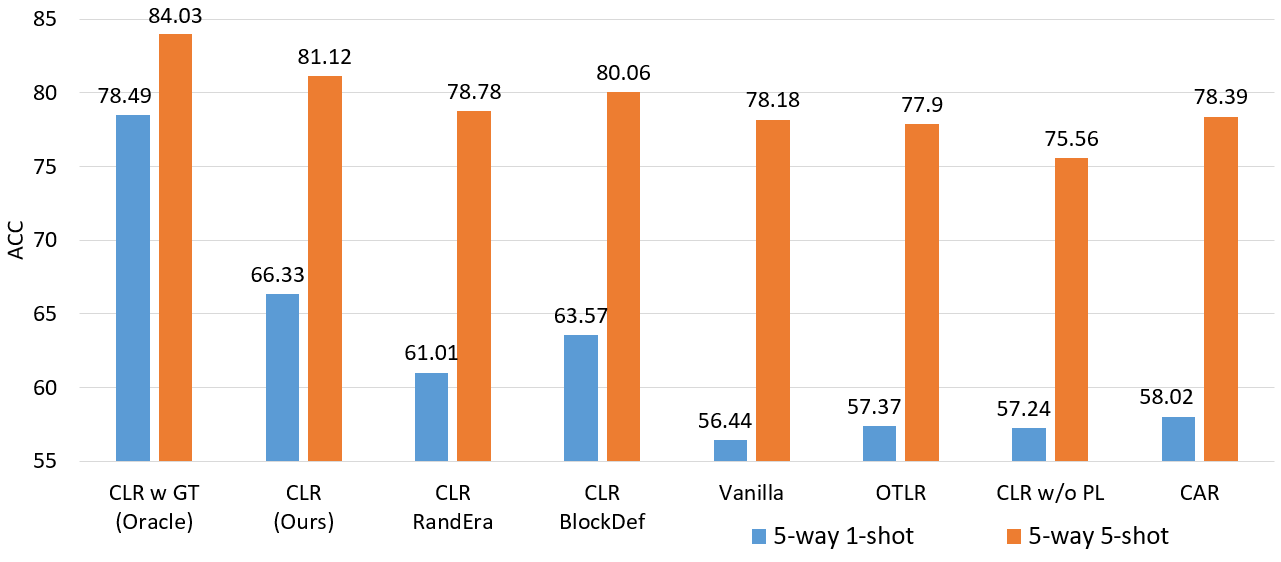}
	\end{tabular}
	\caption{The performance of different variants on MiniImagenet. The proposed CLR is the best on both 1-shot and 5-shot settings. \label{Fig:ablation_study}}
\end{figure}

The evaluation results on MiniImagenet are illustrated in Fig.~\ref{Fig:ablation_study}.
The CLR achieves the best results.
This verifies all components of our approach are important and complementary for the best performance.
The CLR w/o PL performs comparably with Vanilla. 
It is better than Vanilla in 1-shot but slightly worse than Vanilla in 5-shot testing. 
This method randomly select unlabeled images to alter support set. 
Although the image selection could be probably wrong, the classification accuracy still roughly holds.
This indicates the robustness of continual local replacement even when wrong unlabeled images are selected to replace.    
Moreover, the CAR (i.e. original pseudo-labeling method) performs slightly better than Vanilla, but obviously worse than CLR. 
This shows that the CLR is more robust than CAR.
For the CAR method without local replacement, the wrong predictions on unlabeled images could distract the learning objective and limit model's performance.

\subsection{Why Continual Local Replacement Works}
We conduct experiments to quantitatively and qualitatively analyze why CLR works. 

\begin{figure*}[h]
	\centering
	\begin{tabular}{c}
		\includegraphics[width=1.0\textwidth]{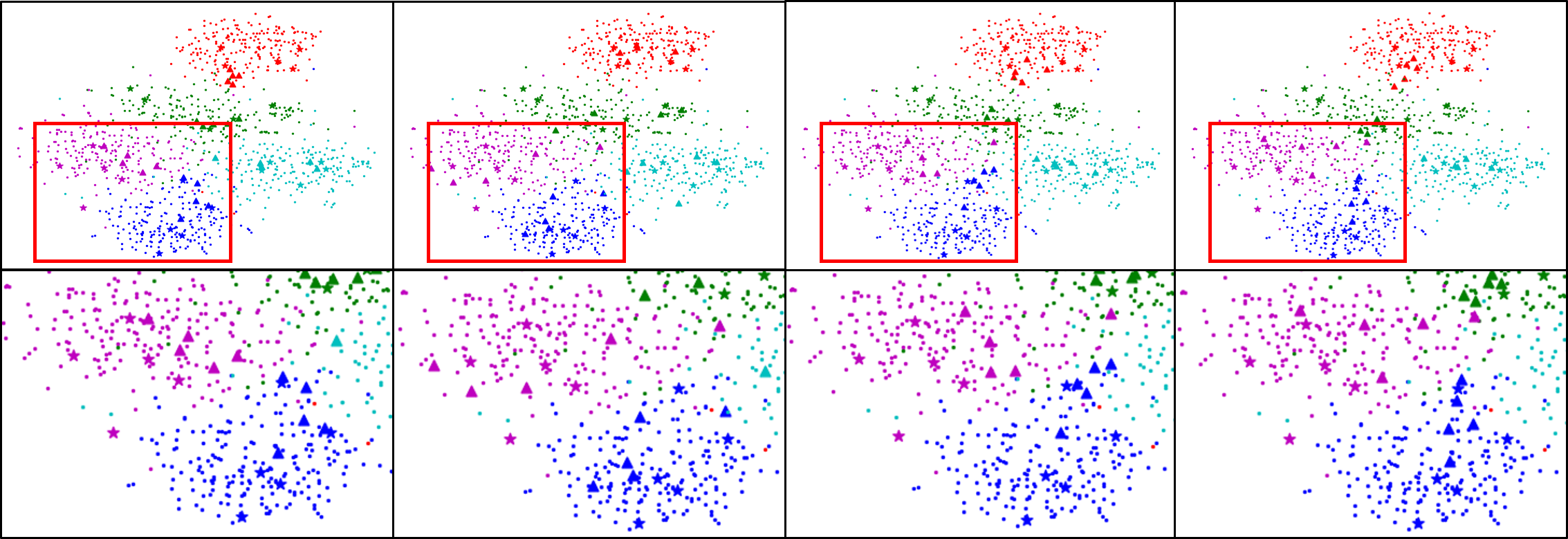}
	\end{tabular}
	\caption{The T-SNE visualization of CLR features on four 5-shot 5-way fine-tune epochs. The embedded features of support set images $d_{support}$ (marked as stars) and query set images $d_{query}$ (marked as dots) are fixed. But the features of local replaced images $d_{support}^s$ (marked as triangles) can change on every fine-tune epoch. It provides a larger semantic space for classifier tuning and thus a better classification decision boundary can be learned. Best viewed in color with zoom. \label{Fig:fine_tune_tsne}}
\end{figure*}

\textbf{CLR introduces plentiful semantic information.} 
On every fine-tune epoch, CLR makes the feature representation of locally replaced images $d_{support}^s$ vary in their embedded space.
This provides a larger semantic space for classifier tuning. Hence, a better classification decision boundary can be learned. 
As illustrated in Fig.~\ref{Fig:fine_tune_tsne}, four fine-tune epochs are showed with T-SNE visualization~\cite{maaten2008visualizing}.
The stars denote the features of $d_{support}$, triangles represent $d_{support}^s$ and other small dots are query images $d_{query}$.
In all fine-tune epochs, the positions of stars and dots are always fixed since we don't update feature extractor backbone during fine tune stage.
With continual local replacement, however, the positions of triangles can move dramatically inside its semantic clusters (see the zoom-in figures in the second row). 
Similar phenomenons can also be observed in the other three semantic clusters (i.e. the red, green and cyan clusters). 

\begin{figure}[h]
	\centering
	\begin{tabular}{c}
		\includegraphics[width=0.47\textwidth]{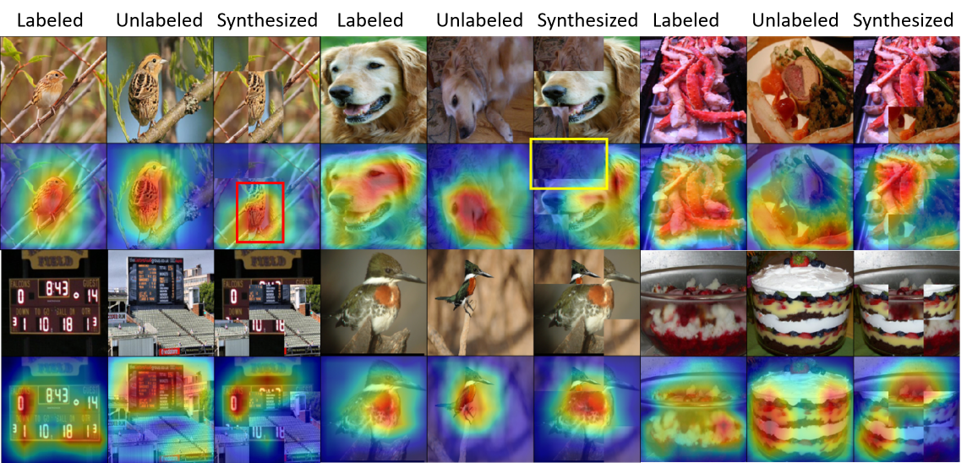}
	\end{tabular}
	\caption{The saliency maps on labeled, unlabeled and synthesized images. The local replacement can be either interpreted as new semantic information (indicated in red rectangle) or partial occlusion (indicated in yellow rectangle) from the regions of unlabeled image. Best viewed in color with zoom. \label{Fig:saliency_map}}
\end{figure}

In addition, the saliency maps are shown in Fig.~\ref{Fig:saliency_map}. 
The local replacement can be either interpreted as new semantic information which is indicated in top-left red rectangle, or partial occlusion which is illustrated in top-side yellow rectangle.

\begin{figure*}[h]
	\centering
	\begin{tabular}{cccc}
		\includegraphics[width=0.23\textwidth]{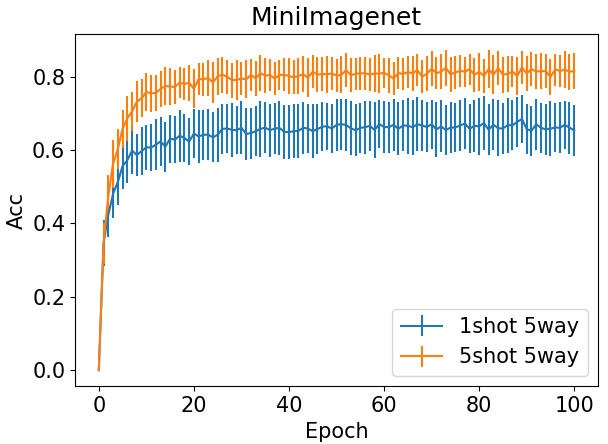}&
		\includegraphics[width=0.23\textwidth]{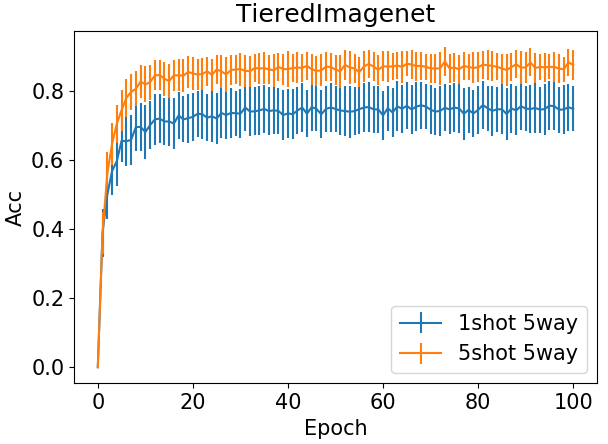}&
		\includegraphics[width=0.23\textwidth]{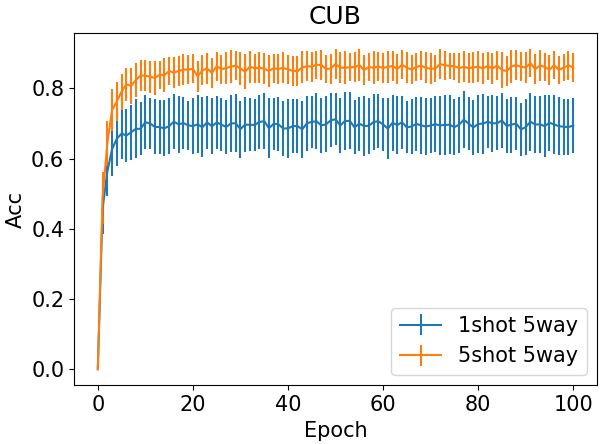}&
		\includegraphics[width=0.23\textwidth]{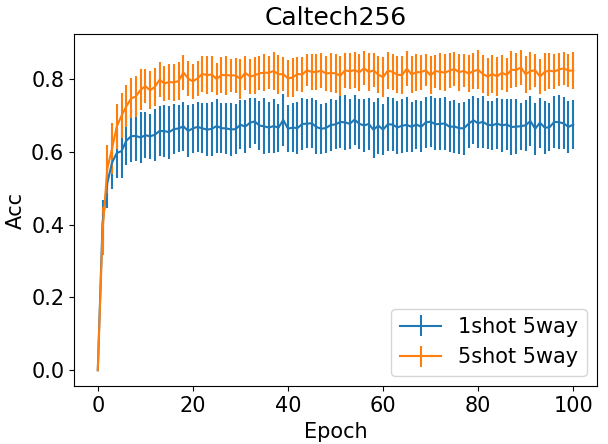}
	\end{tabular}
	\caption{The statistics of classification accuracy change on 100 testing episodes. The error bars indicate the standard deviation and the curves represent the mean values of accuracy. All curves can converge quickly.  \label{Fig:epoch_acc_curve}}
\end{figure*}


\textbf{Robustness of CLR.}
Fig.~\ref{Fig:epoch_acc_curve} shows the accuracy change statistics with the 100 fine-tune iteration steps in 100 randomly episodic testing cases. 
The error bars indicate the standard deviation and the curves represent the mean values of accuracy. 
Although the overall content of $d_{support}^s$ is changed during the fine tune, it always contains stable semantic information from original support set.
Hence, the fine tune procedures can converge quickly in all experiments.
Besides, we notice that the standard deviations in 1-shot are usually larger than 5-shot setting.
This may be because the classifier can learn a more stable and reliable initialization when it is with more labeled images in 5-shot setting.

Moreover, we also evaluate CLR on high-way few-shot testing.
The results are showed in Table~\ref{Tab:high_way_miniimagenet}.
Our method achieves the best results.
Note that the performance of CLR can be constantly improved with more replaced blocks. 
It still achieves the best trade-off when a maximum of 6 blocks replaced.
This indicates that 
(1) our approach still works well even though the overall accuracy is relatively low on high-way setting which is naturally more difficult than the standard 5-way setting.
(2) Since 6 replaced blocks can always achieve the best results, the number of replaced blocks hyper-parameter is insensitive to different datasets and experimental settings

\begin{table}[h]
	\centering
	\scriptsize
	\caption{20-way testing results on MiniImageNet dataset.  \label{Tab:high_way_miniimagenet}}
	\begin{tabular}{|c|c|cc|}
		\hline
		\multirow{2}{*}{Methods} & \multirow{2}{*}{Arch.} & \multicolumn{2}{c|}{ MiniImageNet (\%) } \\
		& & 20-way 1-shot & 20-way 5-shot  \\
		\hline 
		MatchNet~\cite{vinyals2016matching} & ResNet-18	&  25.30 $\pm$ 0.29 & 36.78$\pm$0.25  \\
		\hline
		ProtoNet~\cite{snell2017prototypical} & ResNet-18 & 26.50 $\pm$ 0.30 & 44.96$\pm$0.26 	\\
		\hline
		RelationNet~\cite{sung2018learning} & ResNet-18	& 23.75 $\pm$ 0.30 & 39.17$\pm$0.25   	\\
		\hline
		Baseline~\cite{chen2019closerfewshot} & ResNet-18 & 24.75 $\pm$ 0.28 & 42.03$\pm$0.25 \\
		\hline
		Baseline++~\cite{chen2019closerfewshot} & ResNet-18	& 25.58 $\pm$ 0.27  & 50.85$\pm$0.25  	\\
		\hline
		CLR replace max 0 block & ResNet-18 & 30.62 $\pm$ 0.30  & 52.18 $\pm$ 0.26   \\
		CLR replace max 3 blocks & ResNet-18 & 32.66 $\pm$ 0.29  & 53.62 $\pm$ 0.25 \\
		CLR replace max 6 blocks (ours) & ResNet-18 & \textbf{34.94 $\pm$ 0.33} & \textbf{55.20 $\pm$ 0.26} \\
		CLR replace max 9 blocks & ResNet-18 & 32.48 $\pm$ 0.36  & 53.62 $\pm$ 0.28  \\
		\hline
	\end{tabular}
\end{table}


\section{Conclusion}
In this paper, we present a continual local replacement method for few-shot image recognition.
It leverages the content of unlabeled images to synthesize new images for training.
To introduce more useful semantic information, a pseudo labeling strategy is applied during the fine-tune stage.
It continually selects semantically similar images to locally replace labeled ones.
Our strategy is simple yet effective on few-shot image recognition.
Extensive experiments show that it can significantly enlarge the capacity of semantic information and achieve new state-of-the-art results.

\bibliographystyle{named}
\bibliography{ref_original}

\begin{thebibliography}{}

\bibitem[\protect\citeauthoryear{Berthelot \bgroup \em et al.\egroup
  }{2019}]{berthelot2019mixmatch}
David Berthelot, Nicholas Carlini, Ian Goodfellow, Nicolas Papernot, Avital
  Oliver, and Colin Raffel.
\newblock Mixmatch: A holistic approach to semi-supervised learning.
\newblock In {\em Advances in neural information processing systems}, 2019.

\bibitem[\protect\citeauthoryear{Cai \bgroup \em et al.\egroup
  }{2018}]{cai2018memory}
Qi~Cai, Yingwei Pan, Ting Yao, Chenggang Yan, and Tao Mei.
\newblock Memory matching networks for one-shot image recognition.
\newblock In {\em Proceedings of the IEEE Conference on Computer Vision and
  Pattern Recognition}, pages 4080--4088, 2018.

\bibitem[\protect\citeauthoryear{Chen \bgroup \em et al.\egroup
  }{2019a}]{chen2019closerfewshot}
Wei-Yu Chen, Yen-Cheng Liu, Zsolt Kira, Yu-Chiang Wang, and Jia-Bin Huang.
\newblock A closer look at few-shot classification.
\newblock In {\em International Conference on Learning Representations}, 2019.

\bibitem[\protect\citeauthoryear{Chen \bgroup \em et al.\egroup
  }{2019b}]{chen2019image}
Zitian Chen, Yanwei Fu, Kaiyu Chen, and Yu-Gang Jiang.
\newblock Image block augmentation for one-shot learning.
\newblock In {\em Proceedings of the AAAI Conference on Artificial
  Intelligence}, volume~33, pages 3379--3386, 2019.

\bibitem[\protect\citeauthoryear{Chen \bgroup \em et al.\egroup
  }{2019c}]{chen2019image_mixup}
Zitian Chen, Yanwei Fu, Yu-Xiong Wang, Lin Ma, Wei Liu, and Martial Hebert.
\newblock Image deformation meta-networks for one-shot learning.
\newblock In {\em Proceedings of the IEEE Conference on Computer Vision and
  Pattern Recognition}, pages 8680--8689, 2019.

\bibitem[\protect\citeauthoryear{Chen \bgroup \em et al.\egroup
  }{2019d}]{cheny2019multi}
Zitian Chen, Yanwei Fu, Yinda Zhang, Yu-Gang Jiang, Xiangyang Xue, and Leonid
  Sigal.
\newblock Multi-level semantic feature augmentation for one-shot learning.
\newblock {\em IEEE Transactions on Image Processing}, 2019.

\bibitem[\protect\citeauthoryear{Dhillon \bgroup \em et al.\egroup
  }{2020}]{dhillon2020baseline}
Guneet~S Dhillon, Pratik Chaudhari, Avinash Ravichandran, and Stefano Soatto.
\newblock A baseline for few-shot image classification.
\newblock In {\em International Conference on Learning Representations}, 2020.

\bibitem[\protect\citeauthoryear{Fei-Fei \bgroup \em et al.\egroup
  }{2006}]{fei2006one}
Li~Fei-Fei, Rob Fergus, and Pietro Perona.
\newblock One-shot learning of object categories.
\newblock {\em IEEE transactions on pattern analysis and machine intelligence},
  28(4):594--611, 2006.

\bibitem[\protect\citeauthoryear{Finn \bgroup \em et al.\egroup
  }{2017}]{finn2017model}
Chelsea Finn, Pieter Abbeel, and Sergey Levine.
\newblock Model-agnostic meta-learning for fast adaptation of deep networks.
\newblock In {\em Proceedings of the 34th International Conference on Machine
  Learning-Volume 70}, pages 1126--1135. JMLR. org, 2017.

\bibitem[\protect\citeauthoryear{Garcia and Bruna}{2018}]{garcia2018few}
Victor Garcia and Joan Bruna.
\newblock Few-shot learning with graph neural networks.
\newblock In {\em International Conference on Learning Representations}, 2018.

\bibitem[\protect\citeauthoryear{Gidaris \bgroup \em et al.\egroup
  }{2019}]{gidaris2019boosting}
Spyros Gidaris, Andrei Bursuc, Nikos Komodakis, Patrick P{\'e}rez, and Matthieu
  Cord.
\newblock Boosting few-shot visual learning with self-supervision.
\newblock In {\em Proceedings of the IEEE International Conference on Computer
  Vision}, pages 8059--8068, 2019.

\bibitem[\protect\citeauthoryear{Grandvalet and
  Bengio}{2005}]{grandvalet2005semi}
Yves Grandvalet and Yoshua Bengio.
\newblock Semi-supervised learning by entropy minimization.
\newblock In {\em Advances in neural information processing systems}, pages
  529--536, 2005.

\bibitem[\protect\citeauthoryear{Griffin \bgroup \em et al.\egroup
  }{2007}]{griffin2007caltech}
Gregory Griffin, Alex Holub, and Pietro Perona.
\newblock Caltech-256 object category dataset.
\newblock In {\em California Institute of Technology}, 2007.

\bibitem[\protect\citeauthoryear{Hariharan and
  Girshick}{2017}]{hariharan2017low}
Bharath Hariharan and Ross Girshick.
\newblock Low-shot visual recognition by shrinking and hallucinating features.
\newblock In {\em Proceedings of the IEEE International Conference on Computer
  Vision}, pages 3018--3027, 2017.

\bibitem[\protect\citeauthoryear{He \bgroup \em et al.\egroup
  }{2016}]{he2016deep}
Kaiming He, Xiangyu Zhang, Shaoqing Ren, and Jian Sun.
\newblock Deep residual learning for image recognition.
\newblock In {\em Proceedings of the IEEE conference on computer vision and
  pattern recognition}, pages 770--778, 2016.

\bibitem[\protect\citeauthoryear{Koch \bgroup \em et al.\egroup
  }{2015}]{koch2015siamese}
Gregory Koch, Richard Zemel, and Ruslan Salakhutdinov.
\newblock Siamese neural networks for one-shot image recognition.
\newblock In {\em ICML deep learning workshop}, volume~2, 2015.

\bibitem[\protect\citeauthoryear{Kornblith \bgroup \em et al.\egroup
  }{2019}]{kornblith2019better}
Simon Kornblith, Jonathon Shlens, and Quoc~V Le.
\newblock Do better imagenet models transfer better?
\newblock In {\em Proceedings of the IEEE Conference on Computer Vision and
  Pattern Recognition}, pages 2661--2671, 2019.

\bibitem[\protect\citeauthoryear{Krizhevsky \bgroup \em et al.\egroup
  }{2012}]{krizhevsky2012imagenet}
Alex Krizhevsky, Ilya Sutskever, and Geoffrey~E Hinton.
\newblock Imagenet classification with deep convolutional neural networks.
\newblock In {\em Advances in neural information processing systems}, pages
  1097--1105, 2012.

\bibitem[\protect\citeauthoryear{Lake \bgroup \em et al.\egroup
  }{2011}]{lake2011one}
Brenden Lake, Ruslan Salakhutdinov, Jason Gross, and Joshua Tenenbaum.
\newblock One shot learning of simple visual concepts.
\newblock In {\em Proceedings of the annual meeting of the cognitive science
  society}, volume~33, 2011.

\bibitem[\protect\citeauthoryear{Lee \bgroup \em et al.\egroup
  }{2019}]{lee2019meta}
Kwonjoon Lee, Subhransu Maji, Avinash Ravichandran, and Stefano Soatto.
\newblock Meta-learning with differentiable convex optimization.
\newblock In {\em Proceedings of the IEEE Conference on Computer Vision and
  Pattern Recognition}, pages 10657--10665, 2019.

\bibitem[\protect\citeauthoryear{Lee}{2013}]{lee2013pseudo}
Dong-Hyun Lee.
\newblock Pseudo-label: The simple and efficient semi-supervised learning
  method for deep neural networks.
\newblock In {\em Workshop on Challenges in Representation Learning, ICML},
  volume~3, page~2, 2013.

\bibitem[\protect\citeauthoryear{Liu \bgroup \em et al.\egroup
  }{2019}]{liu2019learning}
Yanbin Liu, Juho Lee, Minseop Park, Saehoon Kim, Eunho Yang, Sung~Ju Hwang, and
  Yi~Yang.
\newblock Learning to propagate labels: Transductive propagation network for
  few-shot learning.
\newblock In {\em International Conference on Learning Representations}, 2019.

\bibitem[\protect\citeauthoryear{Maaten and
  Hinton}{2008}]{maaten2008visualizing}
Laurens van~der Maaten and Geoffrey Hinton.
\newblock Visualizing data using t-sne.
\newblock {\em Journal of machine learning research}, 9(Nov):2579--2605, 2008.

\bibitem[\protect\citeauthoryear{Miyato \bgroup \em et al.\egroup
  }{2018}]{miyato2018virtual}
Takeru Miyato, Shin-ichi Maeda, Masanori Koyama, and Shin Ishii.
\newblock Virtual adversarial training: a regularization method for supervised
  and semi-supervised learning.
\newblock {\em IEEE transactions on pattern analysis and machine intelligence},
  41(8):1979--1993, 2018.

\bibitem[\protect\citeauthoryear{Qi \bgroup \em et al.\egroup
  }{2018}]{qi2018low}
Hang Qi, Matthew Brown, and David~G Lowe.
\newblock Low-shot learning with imprinted weights.
\newblock In {\em Proceedings of the IEEE Conference on Computer Vision and
  Pattern Recognition}, pages 5822--5830, 2018.

\bibitem[\protect\citeauthoryear{Ravi and
  Larochelle}{2017}]{ravi2017optimization}
Sachin Ravi and Hugo Larochelle.
\newblock Optimization as a model for few-shot learning.
\newblock In {\em International Conference on Learning Representations}, 2017.

\bibitem[\protect\citeauthoryear{Ren \bgroup \em et al.\egroup
  }{2018}]{ren2018meta}
Mengye Ren, Eleni Triantafillou, Sachin Ravi, Jake Snell, Kevin Swersky,
  Joshua~B Tenenbaum, Hugo Larochelle, and Richard~S Zemel.
\newblock Meta-learning for semi-supervised few-shot classification.
\newblock In {\em International Conference on Learning Representations}, 2018.

\bibitem[\protect\citeauthoryear{Rusu \bgroup \em et al.\egroup
  }{2019}]{rusu2019meta}
Andrei~A Rusu, Dushyant Rao, Jakub Sygnowski, Oriol Vinyals, Razvan Pascanu,
  Simon Osindero, and Raia Hadsell.
\newblock Meta-learning with latent embedding optimization.
\newblock In {\em International Conference on Learning Representations}, 2019.

\bibitem[\protect\citeauthoryear{Sajjadi \bgroup \em et al.\egroup
  }{2016}]{sajjadi2016regularization}
Mehdi Sajjadi, Mehran Javanmardi, and Tolga Tasdizen.
\newblock Regularization with stochastic transformations and perturbations for
  deep semi-supervised learning.
\newblock In {\em Advances in Neural Information Processing Systems}, pages
  1163--1171, 2016.

\bibitem[\protect\citeauthoryear{Salakhutdinov \bgroup \em et al.\egroup
  }{2012}]{salakhutdinov2012one}
Ruslan Salakhutdinov, Joshua Tenenbaum, and Antonio Torralba.
\newblock One-shot learning with a hierarchical nonparametric bayesian model.
\newblock In {\em Proceedings of ICML Workshop on Unsupervised and Transfer
  Learning}, pages 195--206, 2012.

\bibitem[\protect\citeauthoryear{Santoro \bgroup \em et al.\egroup
  }{2016}]{santoro2016one}
Adam Santoro, Sergey Bartunov, Matthew Botvinick, Daan Wierstra, and Timothy
  Lillicrap.
\newblock One-shot learning with memory-augmented neural networks.
\newblock {\em arXiv preprint arXiv:1605.06065}, 2016.

\bibitem[\protect\citeauthoryear{Simonyan and
  Zisserman}{2014}]{simonyan2014very}
Karen Simonyan and Andrew Zisserman.
\newblock Very deep convolutional networks for large-scale image recognition.
\newblock {\em arXiv preprint arXiv:1409.1556}, 2014.

\bibitem[\protect\citeauthoryear{Snell \bgroup \em et al.\egroup
  }{2017}]{snell2017prototypical}
Jake Snell, Kevin Swersky, and Richard Zemel.
\newblock Prototypical networks for few-shot learning.
\newblock In {\em Advances in Neural Information Processing Systems}, pages
  4077--4087, 2017.

\bibitem[\protect\citeauthoryear{Sun \bgroup \em et al.\egroup
  }{2019}]{sun2019meta}
Qianru Sun, Yaoyao Liu, Tat-Seng Chua, and Bernt Schiele.
\newblock Meta-transfer learning for few-shot learning.
\newblock In {\em Proceedings of the IEEE Conference on Computer Vision and
  Pattern Recognition}, pages 403--412, 2019.

\bibitem[\protect\citeauthoryear{Sung \bgroup \em et al.\egroup
  }{2018}]{sung2018learning}
Flood Sung, Yongxin Yang, Li~Zhang, Tao Xiang, Philip~HS Torr, and Timothy~M
  Hospedales.
\newblock Learning to compare: Relation network for few-shot learning.
\newblock In {\em Proceedings of the IEEE Conference on Computer Vision and
  Pattern Recognition}, pages 1199--1208, 2018.

\bibitem[\protect\citeauthoryear{Vinyals \bgroup \em et al.\egroup
  }{2016}]{vinyals2016matching}
Oriol Vinyals, Charles Blundell, Timothy Lillicrap, Daan Wierstra, et~al.
\newblock Matching networks for one shot learning.
\newblock In {\em Advances in neural information processing systems}, pages
  3630--3638, 2016.

\bibitem[\protect\citeauthoryear{Wah \bgroup \em et al.\egroup
  }{2011}]{wah2011caltech}
Catherine Wah, Steve Branson, Peter Welinder, Pietro Perona, and Serge
  Belongie.
\newblock The caltech-ucsd birds-200-2011 dataset.
\newblock In {\em California Institute of Technology}, 2011.

\bibitem[\protect\citeauthoryear{Wang \bgroup \em et al.\egroup
  }{2017}]{wang2017multi}
Peng Wang, Lingqiao Liu, Chunhua Shen, Zi~Huang, Anton van~den Hengel, and Heng
  Tao~Shen.
\newblock Multi-attention network for one shot learning.
\newblock In {\em Proceedings of the IEEE Conference on Computer Vision and
  Pattern Recognition}, pages 2721--2729, 2017.

\bibitem[\protect\citeauthoryear{Wang \bgroup \em et al.\egroup
  }{2018}]{wang2018low}
Yu-Xiong Wang, Ross Girshick, Martial Hebert, and Bharath Hariharan.
\newblock Low-shot learning from imaginary data.
\newblock In {\em Proceedings of the IEEE Conference on Computer Vision and
  Pattern Recognition}, pages 7278--7286, 2018.

\bibitem[\protect\citeauthoryear{Zhong \bgroup \em et al.\egroup
  }{2020}]{zhong2020random}
Zhun Zhong, Liang Zheng, Guoliang Kang, Shaozi Li, and Yi~Yang.
\newblock Random erasing data augmentation.
\newblock In {\em Proceedings of the AAAI Conference on Artificial Intelligence
  (AAAI)}, 2020.

\bibitem[\protect\citeauthoryear{Zhou \bgroup \em et al.\egroup
  }{2018}]{zhou2018deep}
Fengwei Zhou, Bin Wu, and Zhenguo Li.
\newblock Deep meta-learning: Learning to learn in the concept space.
\newblock {\em arXiv preprint arXiv:1802.03596}, 2018.

\end{thebibliography}

\end{document}